\documentclass[manuscript]{IEEEtran}
\IEEEoverridecommandlockouts
\usepackage{cite}
\usepackage{amsmath,amssymb,amsfonts}
\usepackage{algorithmic}
\usepackage{graphicx}
\usepackage{textcomp}
\usepackage{xcolor}
\usepackage{tabularx}
\usepackage{comment}
\def\BibTeX{{\rm B\kern-.05em{\sc i\kern-.025em b}\kern-.08em
    T\kern-.1667em\lower.7ex\hbox{E}\kern-.125emX}}

\usepackage{todonotes}
\usepackage{flushend}
\begin{document}

\title{The Wilhelm Tell Dataset of Affordance Demonstrations\\
\thanks{
        
            This work was funded by the FET-Open Project \#951846 ``MUHAI – Meaning and Understanding for Human-centric AI'' by the EU Pathfinder and Horizon 2020 Program.
           This work has also been supported by the German Research Foundation DFG, as part of Collaborative Research Center 1320 Project-ID 329551904 ``EASE -- Everyday Activity Science and Engineering'', University of Bremen (https://www.ease-crc.org).
            The research was conducted in subproject ``P01 – Embodied semantics for the language of action and change: Combining analysis, reasoning and simulation''.
            }
}



\author{\IEEEauthorblockN{Rachel Ringe}
\IEEEauthorblockA{\textit{Digital Media Lab} \\
\textit{University of Bremen}\\
Bremen, Germany \\
rringe@uni-bremen.de
}

\IEEEauthorblockN{Mihai Pomarlan}
\IEEEauthorblockA{\textit{Applied Linguistics} \\ \textit{University of Bremen} \\
Bremen, Germany \\
pomarlan@uni-bremen.de
}

\IEEEauthorblockN{Nikolaos Tsiogkas}
\IEEEauthorblockA{\textit{Department of Computer Science} \\
\textit{KU Leuven}\\
{Leuven, Belgium} \\
nikolaos.tsiogkas@kuleuven.be
}

\IEEEauthorblockN{Stefano De Giorgis}
\IEEEauthorblockA{\textit{Institute of Cognitive Sciences and Technologies} \\ \textit{National Research Council} \\ Catania, Italy \\
stefano.degiorgis@cnr.it}

\IEEEauthorblockN{Maria Hedblom}
\IEEEauthorblockA{\textit{Department of Computing} \\\textit{J{\"o}nk{\"o}ping School of Engineering} \\ J{\"o}nk{\"o}ping, Sweden \\
maria.hedblom@ju.se}

\IEEEauthorblockN{Rainer Malaka}
\IEEEauthorblockA{\textit{Digital Media Lab} \\
\textit{University of Bremen}\\
Bremen, Germany \\malaka@tzi.de
}
}
\maketitle

\begin{abstract}
Affordances -- i.e. possibilities for action that an environment or objects in it provide -- are important for robots operating in human environments to perceive. Existing approaches train such capabilities on annotated static images or shapes. This work presents a novel dataset for affordance learning of common household tasks. Unlike previous approaches, our dataset consists of video sequences demonstrating the tasks from first- and third-person perspectives, along with metadata about the affordances that are manifested in the task, and is aimed towards training perception systems to recognize affordance manifestations. The demonstrations were collected from several participants and in total record about seven hours of human activity. The variety of task performances also allows studying preparatory maneuvers that people may perform for a task, such as how they arrange their task space, which is also relevant for collaborative service robots.
\end{abstract}

\begin{IEEEkeywords}
affordance demonstrations, affordance recognition, domestic service robotics
\end{IEEEkeywords}

\section{Introduction}


The general goal of robotics 
is to replace humans in work that is dangerous, repetitive, or requiring high levels of precision in combination with high repeatability and production speed. Despite this goal, most robotic executed tasks are pre-programmed sequences of motions, given as an input from a human 
expert, making the general adoption of robotics technolgies in every-day life close to impossible. To remove this barrier of general adoption, robots should be able to learn in a multi-modal way, as humans do, which includes verbal or textual instructions~\cite{mericcli2014interactive}, demonstartion~\cite{ravichandar2020recent}, and observation~\cite{ikeuchi2018describing}.

This work focuses on enabling the development of methods that will allow robots to learn while observing humans performing specific tasks. Observation is the least disruptive method, as it allows the human to perform a task uninterrupted. 
Direct observation where the robot is actually deployed presents the ideal case, methods that are fast, accurate, and robust enough, are required. Despite the availability of methods in the literature, various shortcomings are not allowing their use out of laboratory setups \cite{lu2020service}. To allow the developement of more advanced methods, more observation data is needed, so that the scientific community can refine and test their approaches. This work presents a curated set of such data that can be used for robot training purposes.

The rest of the paper is organized as follows. Section \ref{lbl:background} presents relevant background. Section \ref{lbl:overview} gives an overview of the study documented in this work. Section \ref{lbl:methods} details the methods used. Section \ref{lbl:dataset} describes the dataset. Section \ref{lbl:usage} provides information regarding the use of the presented 
data. Finally, in section \ref{lbl:discussion}, a discussion is provided and relevant tracks of future work that use the dataset are layed out.

\section{Background}
\label{lbl:background}

Taking inspiration from generic manipulation \cite{khazatsky2024droid} and care-focused activity recognition \cite{baselizadeh2024prima} datasets, the presented dataset focuses on affordance and activity recognition in common household settings. It fills the void in this domain between datasets for object detection and recognition \cite{ishida2020semi,lv2021object, azagra2017multimodal}, social behaviour \cite{ko2021air}, voice recognition \cite{brinckhaus2021robocup}, generic human activity \cite{wang2023dataset, jang2020etri}, social navigation \cite{karnan2022socially}, and human-robot interaction \cite{bu2024ssup}. 


The notion of affordance is due to Gibson, originally defined as ``what an environment provides to an agent, for good or for ill''(\cite{gibson1979ecological}, pg. 127). Several formal treatments of affordances have been proposed in the knowledge representation and ontology engineering literature~\cite{bessler2020,moralez2016affordance,toyoshima2018modeling}. The notion of affordance is also important in interface design~\cite{karat2000,Masoudi_Fadel_Pagano_Elena_2019}. The detailed definition of what an affordance is therefore differs depending on field and even author, however, a common element is that affordances are, or at least relate to, possibilities for action.

The field of robotics has also displayed an interest in affordances in recent years~\cite{affpersurvey}. In this literature, ``affordance'' often means, or is associated with, parts of an object that are relevant for a particular task. E.g. the handle of a mug would be a grasping affordance. Datasets of annotated shapes such as AfNet~\cite{AfNet} and AffordanceNet~\cite{AffordanceNet} have been published to assist in the development and evaluation of machine learning solutions to recognize affordances, and several such systems have been described~\cite{Nguyen2023open,Luo2022LearningAG,Chen2021CerberusTJ,Peng2022OpenScene3S}.

Related to common household activities, the works of \cite{mur2023multi,luo2023learning} present methods to automatically detect affordances based on video inputs. They use datasets such as the ones presented in \cite{fang2018demo2vec,damen2020epic}, to train NN-based architectures that perform the afordance learning and grounding. Compared to \cite{fang2018demo2vec}, this work covers a larger set of affordances, particularly focusing on affordances that involve the interaction of multiple objects with each other and the agent as opposed to interactions between an agent and an object. Regarding the work presented in \cite{damen2020epic} and \cite{grauman2022ego4d}, our work provides an extra perspective, other than the ego-perspective, similar to an observer of a demonstration. Finally, the work presented in \cite{bahl2023affordances}, presents another relevant affordance detection system, but unfortunatelly the used dataset is not publically available for the time being to the best of the authors' knowledge.

Apart from approaches that learn affordances from video~\cite{mur2023multi,luo2023learning}, many of the systems for affordance detection cited above ultimately rely on datasets of annotated \emph{shapes} or \emph{images} to specify affordances\footnote{This dependency also exists in ``annotation free'' methods such as~\cite{Peng2022OpenScene3S}, where it is hidden in the training data of the component multimodal (image and captions) network CLIP.}. This does not, in our opinion, robustly capture the idea of affordance as being, or being associated with, possibilities for action. Rather, an agent should learn to recognize an affordance by \emph{observing events that manifest it}: not to recognize a handle because it happens to be in some set of arbitrary shapes, but to recognize it as a handle because it has seen a similar shape being grabbed. Therefore, 
we focused not on shape annotations, but rather in collecting demonstrations of affordances via episodes in which people performed simple tasks.

\section{Study Overview}
\label{lbl:overview}

During the study participants were equipped with a GoPro headset camera and then asked to stand in front of a table with a variety of household objects and instructed to perform various small tasks centered around apples, using the objects to demonstrate affordances. The resulting videos can then be used to train vision systems to recognize activities and the functional parts of objects that are relevant for particular tasks.

The affordances considered in the study mostly come from AffordanceNet~\cite{AffordanceNet} and AfNet\cite{AfNet}, which are pre-existing datasets of shapes with parts annotated to highlight regions on an object that are important for a particular affordance. We understood the presence of an affordance in a preexisting dataset to be an indication that the affordance is considered important for robotics applications. However, our dataset does not include demonstrations for all AfNet and AffordanceNet affordances. Reasons for exclusion refer to aspects such as the involvement of sound, or the affordance requiring no demonstration involving a human at all. We wished however to focus on affordances which can be visually demonstrated by people. Conversely, we added two affordances not present in AfNet/AffordanceNet -- Stab and Pour -- because they seemed relevant to us and easy to demonstrate.

We deliberately chose to only give participants a broad indication of what to do, and left at their discretion the manner in which to do it. Thus, different people have performed different preparation maneuvers and item placements, had different ways to perform the tasks and different ways to leave the items when the task was finished.




\section{Methods}
\label{lbl:methods}
\subsection{Activities/Tasks/Affordances}
\begin{figure}[htbp]
    \centering
    \includegraphics[width=\columnwidth]{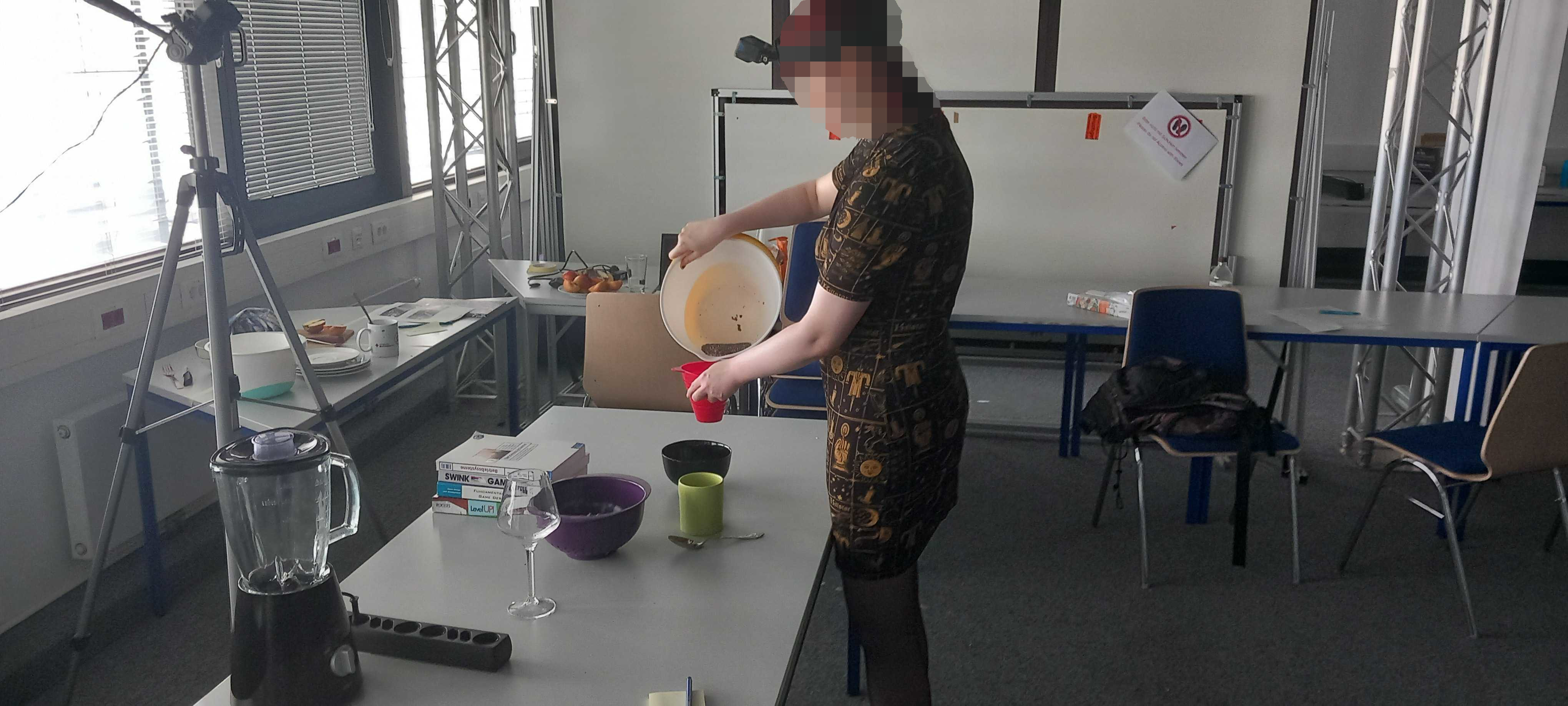}
    \caption{Picture of lab environment showing the table with the objects as well as the recording setup}
    \label{fig:setup}
\end{figure}

The primary aim of the dataset was to show a variety of affordances using everyday objects from multiple perspectives - the ego perspective recorded with a GoPro attached to a headstreap and a frontal perspective one might encounter when receiving a demonstration. This setup can be seen in Fig. \ref{fig:setup}.


Participants received instructions from a researcher sitting at a table to their left. Participants were asked to wait for the researcher to finish an instruction until taking action and to attempt to look at their hands while completing the tasks to improve the quality of the recordings from the camera attached to the headstrap. The researcher would wait for the participant to complete the task and then continue with the next instruction. When new objects were needed, the researcher informed the participant that the objects were being replaced before continuing. In cases where participants were unsure what object was mentioned in the instruction due to language problems, the researcher pointed out the corrent object.

The data was collected in three sessions. In the first session eight participants were asked to complete 14 tasks in two bundles of 7 as shown in Table \ref{table:study1}. Task 3 and 4 were varied with the \texttt{a} option used in the first round and the \texttt{b} option used in the second round of the study. The tools used were switched after the first 7 tasks to provide more variety in the recordings and the apple was replaced with an intact one.

\begin{table}[]
\begin{tabularx}{\linewidth}{|l|X|X|}
\hline
\textbf{} & \textbf{Task Given} & \textbf{Demonstrated Affordance} \\ \hline
1 & Please stab the apple with the fork and lift it. Then move it around a bit and take it off the fork, if needed with your fingers. & Stab 
\\ \hline
2 & Now cut the apple in halves with the knife. & Cut (AffordanceNet) \\ \hline
3 & Please now stab one of the apple halves with the fork and lift it. Then move it around a bit and take it off the fork, if needed with your fingers, and \{(a) return it to the cutting board/(b) put it into the mug\}. & Stab 
\\ \hline
4 & Please put \{(a) the apple pieces/(b) the other apple half\} into the mug. & Contain - ability (AfNet) / Contain (AffordanceNet)\\ \hline
5 & Pick up the mug and move it around a bit, tilting it a bit without spilling its contents. Look into the cup as you move it around. &  Contain - ability (AfNet) / Contain (AffordanceNet) \\ \hline
6 & Now pour the apple halves into the bowl. & Pour 
\\ \hline
7 & Pick up the bowl and move it around a bit, tilting it a bit without spilling its contents. Look into the bowl as you move it around. & Contain - ability (AfNet) / Contain (AffordanceNet) \\ \hline
\end{tabularx}
\caption{Tasks given to participants in the first session of the study. Some tasks have different variants (a), (b).}
\label{table:study1}
\end{table}

For the second and third recording session the list of tasks was expanded with more tasks that demonstrate additional affordances. Due to the time needed for those added tasks they were split into four parts. During the first round participants completed tasks 8-10 in the \texttt{a}-variant and then continued with tasks 1-7 in the \texttt{a}-variant. The objects were then replaced with different variations of the same object and the participants repeated tasks 8-10 and then 1-7 in the \texttt{b}-variant. Afterwards the objects on the table were replaced with different objects and the participants received instructions for tasks 11-20 in the \texttt{a}-variant. Following this, the objects were again replaced with different variations, and participants then instructed to complete tasks 11-20 in the \texttt{b}-variant.

\begin{table}[]
\begin{tabularx}{\linewidth}{|l|X|X|}
\hline
\textbf{} & \textbf{Task Given} & \textbf{Demonstrated Affordance} \\ \hline
8 & Please roll the apple over the table \{(a) with your hand/(b) from hand to hand\} a few times. & 3D - roll ability (AfNet)\\ \hline
9 & Please wrap the apple in the \{(a) dishtowel/(b) tinfoil\}, pick up the wrapped apple, move it around a bit and then unwrap it and return both items to the table. & Wrap - ability (AfNet) \\ \hline
10 & Please take the knife and carve an X into the apple. & Engrave + Incision ability (AfNet)\\ \hline
11 & Please stack the four \{(a) plates/(b) books\} on top of each other. & Stack - ability (AfNet) \\ \hline
12 & Please take the wine glass and roll it over the table \{(a) with your hand/(b) from hand to hand\} a few times. & 2D - roll ability (AfNet) \\ \hline
13 & Please pick up the stack of \{(a)plates/(b) books\} and move it around a bit, then return it to the table. & Stack - ability (AfNet)\\ \hline
14 & Please take the pencil and write the word “apple” \{(a) into the notebook/(b) onto the post-it\}. & Engrave (AfNet) \\ \hline
\end{tabularx}
\caption{Additional Tasks given to participants in the second and third session of the study. Some have different variants (a), (b).}
\label{table:study2}
\end{table}

\begin{table}[]
\begin{tabularx}{\linewidth}{|l|X|X|}
\hline
\textbf{} & \textbf{Task Given} & \textbf{Demonstrated Affordance} \\ \hline
15 & Please pick up the large bowl with \{(a) sugar/(b) chia seeds\} and move it around a bit, without spilling its contents, then return it to the table. &  Contain - ability (AfNet) / Contain (AffordanceNet) \\ \hline
16 & Then take the funnel and use it to transfer the \{(a) sugar/(b) chia seeds\} from the large bowl into the mug. Then remove the funnel. & Flow Support ability (AfNet)\\ \hline
17 & Please take the spoon and use it to transfer the \{(a) sugar/(b) chia seeds\} into the small bowl. Use the spoon for the first four or five spoonfuls, afterwards you can pour the rest. & Scoop (AffordanceNet) \\ \hline
18 & Put apple halves into the small bowl with \{(a) sugar/(b) chia seeds\}. & Contain - ability (AfNet) / Contain (AffordanceNet) \\ \hline
19 & Please place the colander over the empty large bowl. Pour the \{(a) sugar/(b) chia seeds\} and the apple into the colander. Then lift the colander and move it over the bowl until the \{(a) sugar/(b) chia seeds\} drops through, then return it to the table. & Filter - ability (AfNet) \\ \hline
20 & Please pick up the \{(a) mixing cup and insert it into the blender/(b) blender plug and plug it into the extension cord. The extension cord is not plugged in, so there is no danger. Then pick up the extension cord and move it around a bit, then return it to the table\}. & Connect - ability (AfNet) \\ \hline
\end{tabularx}
\caption{Additional Tasks given to participants in the second and third session of the study. Some have different variants (a), (b).}
\label{table:study3}
\end{table}

\subsection{Participant Recruitment, Consent, and Demographics}



The study was conducted with 23 participants (11 female, 12 male) with an average age of 31.57. Participants were recruited personally and via social media. 

Before the start of the study participants were informed, via a consent form, about the data collection process, that their faces would not be visible in the pulished recordings, procedures for recording and storage and video data as well as their right to withdraw their consent at any time, future plans for publication of the collected data, the small but possible danger of injury due to the tools involved, and the small but possible risk of personal data leakage. The study was conducted with at least one of the researchers present at all times to supervise the participants. The collected video data was stored locally on university computers as well as GDPR compliant university servers. After the study participants were offered their cut apples as compensation - leftover apples were not wasted but turned into apple pie.




\section{Dataset}
\label{lbl:dataset}

The final dataset consists of 189 videos in MP4 format with a total of 417 minutes of material. For every participant with a complete data collection there are two or four videos from each perspective - each containing one bundle of tasks - as well as one video per perspective combining the smaller videos into one showing the complete set of actions performed by each participant with the pauses between bundles where researchers moved objects removed.
For 7 participants the video collection was incomplete due to technical difficulties with the recording setup or the participant not completing the full set of tasks. These incomplete entries were still included in the dataset since the existing recordings show the demonstrated tasks sufficiently and consent for publication was not withdrawn by the participants who did not complete all tasks.
The Readme file for the dataset containing information as well as a download link to the video material can be found on GitHub\footnote{https://github.com/ease-crc/WilhelmTellDataset}. The dataset is licensed under a CC-BY 4.0 license.
The videos, still images, and annotation files are hosted on university servers dedicated to maintaining artifacts created during our supporting collaborative research center.

\section{Usage Notes}
\label{lbl:usage}

The intended use case for our dataset is the training and evaluation of visual perception systems for affordance recognition. Training affordance detection from video has previously been described in the literature~\cite{mur2023multi,luo2023learning} but our dataset brings together a novel combination of more viewpoints and more affordances that involve multiple objects interacting. The objects appearing in our dataset may themselves present some challenges for tracking, because they tend to occupy small parts of a frame, may have little to no overlap with themselves from frame to frame, will be occluded and in many angles during performance of a task etc.


Video clips corresponding to the bundles of tasks listed in tables~\ref{table:study1}-\ref{table:study3} have been grouped into sets of 102 training and 44 testing videos. The test videos have corresponding CSV files describing time intervals in which objects interact to manifest an affordance. Some of the frames from the test videos have been extracted as still images and annotated with affordance hotspots for objects.

The typical use-case we envision for our dataset is to train semantic segmentation models so that they can detect affordance hotspots. In this use-case, a researcher would select affordances of interest, feed the training and test videos demonstrating the affordances to an automatic affordance hotspot labeling system in the vein of~\cite{mur2023multi}, and then use the automtically generated annotations to train and test image segmentation models. The goal is to teach the segmentation model to recognize functional parts of objects both within situations where the functional parts are used -- where interaction clues show where the affordance hotspot is -- as well as outside situations of use.

The variety of task performances suggests further use cases, such as studying how people organize space during a task execution -- preparatory maneuvers, what they bring close or set far. This can be used to train perception to anticipate such maneuvers and enable the development of collaborative robots.


\section{Discussion}
\label{lbl:discussion}

We have presented a dataset of video demonstrations of affordances for a few basic tasks. To our knowledge, most approaches to affordance detection so far rely on knowledge extracted from annotations of static shapes/images, and we believe our dataset will contribute towards solutions capable to understand affordances as manifested in events. This appears to us a more robust understanding of affordance as something enabling an action, rather than something (more or less) arbitrarily associated with a static image feature.

We are considering expansions of our dataset in the near future such as different settings with more clutter and/or a realistic kitchen environment. We are particularly interested in asking people to do tasks while being watched by a robot and observing whether the task demonstration changes in the presence of something more agent-like playing the role of observer. Recording in a larger, more realistic kitchen environment is also expected to further encourage the appearance of more preparatory maneuvers and illustrations of how humans organize space for their tasks.

\section*{Acknowledgment}
We thank Lars Hurrelbrink for the dataset icon and header image, as well as our study participants.

\bibliographystyle{IEEEtran}
\bibliography{bibliography}

\begin{thebibliography}{10}
\providecommand{\url}[1]{#1}
\csname url@samestyle\endcsname
\providecommand{\newblock}{\relax}
\providecommand{\bibinfo}[2]{#2}
\providecommand{\BIBentrySTDinterwordspacing}{\spaceskip=0pt\relax}
\providecommand{\BIBentryALTinterwordstretchfactor}{4}
\providecommand{\BIBentryALTinterwordspacing}{\spaceskip=\fontdimen2\font plus
\BIBentryALTinterwordstretchfactor\fontdimen3\font minus \fontdimen4\font\relax}
\providecommand{\BIBforeignlanguage}[2]{{%
\expandafter\ifx\csname l@#1\endcsname\relax
\typeout{** WARNING: IEEEtran.bst: No hyphenation pattern has been}%
\typeout{** loaded for the language `#1'. Using the pattern for}%
\typeout{** the default language instead.}%
\else
\language=\csname l@#1\endcsname
\fi
#2}}
\providecommand{\BIBdecl}{\relax}
\BIBdecl

\bibitem{mericcli2014interactive}
C.~Meri{\c{c}}li, S.~D. Klee, J.~Paparian, and M.~Veloso, ``An interactive approach for situated task specification through verbal instructions,'' in \emph{Proceedings of the 2014 international conference on Autonomous agents and multi-agent systems}, 2014, pp. 1069--1076.

\bibitem{ravichandar2020recent}
H.~Ravichandar, A.~S. Polydoros, S.~Chernova, and A.~Billard, ``Recent advances in robot learning from demonstration,'' \emph{Annual review of control, robotics, and autonomous systems}, vol.~3, no.~1, pp. 297--330, 2020.

\bibitem{ikeuchi2018describing}
K.~Ikeuchi, Z.~Ma, Z.~Yan, S.~Kudoh, and M.~Nakamura, ``Describing upper-body motions based on labanotation for learning-from-observation robots,'' \emph{International Journal of Computer Vision}, vol. 126, pp. 1415--1429, 2018.

\bibitem{lu2020service}
V.~N. Lu, J.~Wirtz, W.~H. Kunz, S.~Paluch, T.~Gruber, A.~Martins, and P.~G. Patterson, ``Service robots, customers and service employees: what can we learn from the academic literature and where are the gaps?'' \emph{Journal of Service Theory and Practice}, vol.~30, no.~3, pp. 361--391, 2020.

\bibitem{khazatsky2024droid}
A.~Khazatsky, K.~Pertsch, S.~Nair, A.~Balakrishna, S.~Dasari, S.~Karamcheti, S.~Nasiriany, M.~K. Srirama, L.~Y. Chen, K.~Ellis \emph{et~al.}, ``Droid: A large-scale in-the-wild robot manipulation dataset,'' \emph{arXiv preprint arXiv:2403.12945}, 2024.

\bibitem{baselizadeh2024prima}
A.~Baselizadeh, M.~Z. Uddin, W.~Khaksar, D.~S. Lindblom, and J.~Torresen, ``Prima-care: Privacy-preserving multi-modal dataset for human activity recognition in care robots,'' in \emph{Companion of the 2024 ACM/IEEE International Conference on Human-Robot Interaction}, 2024, pp. 233--237.

\bibitem{ishida2020semi}
Y.~Ishida and H.~Tamukoh, ``Semi-automatic dataset generation for object detection and recognition and its evaluation on domestic service robots,'' \emph{Journal of Robotics and Mechatronics}, vol.~32, no.~1, pp. 245--253, 2020.

\bibitem{lv2021object}
Y.~Lv, Y.~Fang, W.~Chi, G.~Chen, and L.~Sun, ``Object detection for sweeping robots in home scenes (odsr-ihs): a novel benchmark dataset,'' \emph{IEEE Access}, vol.~9, pp. 17\,820--17\,828, 2021.

\bibitem{azagra2017multimodal}
P.~Azagra, F.~Golemo, Y.~Mollard, M.~Lopes, J.~Civera, and A.~C. Murillo, ``A multimodal dataset for object model learning from natural human-robot interaction,'' in \emph{2017 IEEE/RSJ International Conference on Intelligent Robots and Systems (IROS)}.\hskip 1em plus 0.5em minus 0.4em\relax IEEE, 2017, pp. 6134--6141.

\bibitem{ko2021air}
W.-R. Ko, M.~Jang, J.~Lee, and J.~Kim, ``Air-act2act: Human--human interaction dataset for teaching non-verbal social behaviors to robots,'' \emph{The International Journal of Robotics Research}, vol.~40, no. 4-5, pp. 691--697, 2021.

\bibitem{brinckhaus2021robocup}
E.~Brinckhaus, G.~T. Barnech, M.~Etcheverry, and F.~Andrade, ``Robocup@ home: Evaluation of voice recognition systems for domestic service robots and introducing latino dataset,'' in \emph{2021 Latin American Robotics Symposium (LARS), 2021 Brazilian Symposium on Robotics (SBR), and 2021 Workshop on Robotics in Education (WRE)}.\hskip 1em plus 0.5em minus 0.4em\relax IEEE, 2021, pp. 25--29.

\bibitem{wang2023dataset}
J.~Wang, T.~Zhang, X.~Wu, and L.~Zeng, ``A dataset and system for service robot action interaction based on skeleton action recognition,'' in \emph{2023 8th International Conference on Signal and Image Processing (ICSIP)}.\hskip 1em plus 0.5em minus 0.4em\relax IEEE, 2023, pp. 41--48.

\bibitem{jang2020etri}
J.~Jang, D.~Kim, C.~Park, M.~Jang, J.~Lee, and J.~Kim, ``Etri-activity3d: A large-scale rgb-d dataset for robots to recognize daily activities of the elderly,'' in \emph{2020 IEEE/RSJ International Conference on Intelligent Robots and Systems (IROS)}.\hskip 1em plus 0.5em minus 0.4em\relax IEEE, 2020, pp. 10\,990--10\,997.

\bibitem{karnan2022socially}
H.~Karnan, A.~Nair, X.~Xiao, G.~Warnell, S.~Pirk, A.~Toshev, J.~Hart, J.~Biswas, and P.~Stone, ``Socially compliant navigation dataset (scand): A large-scale dataset of demonstrations for social navigation,'' \emph{IEEE Robotics and Automation Letters}, vol.~7, no.~4, pp. 11\,807--11\,814, 2022.

\bibitem{bu2024ssup}
F.~Bu and W.~Ju, ``Ssup-hri: Social signaling in urban public human-robot interaction dataset,'' \emph{arXiv preprint arXiv:2403.10994}, 2024.

\bibitem{gibson1979ecological}
J.~J. Gibson, \emph{The Ecological Approach to Visual Perception}.\hskip 1em plus 0.5em minus 0.4em\relax Psychology Press Classic Editions, 1979.

\bibitem{bessler2020}
D.~Beßler, R.~Porzel, M.~Pomarlan, M.~Beetz, R.~Malaka, and J.~Bateman, ``A formal model of affordances for flexible robotic task execution,'' in \emph{Proceedings of the 24th European Conference on Artificial Intelligence (ECAI)}, 09 2020.

\bibitem{moralez2016affordance}
L.~A. Moralez, ``Affordance ontology: towards a unified description of affordances as events,'' \emph{Res. Cogitans}, vol.~7, no.~1, pp. 35--45, 2016.

\bibitem{toyoshima2018modeling}
F.~TOYOSHIMA, ``Modeling affordances with dispositions,'' in \emph{Proceedings of the Joint Ontology Workshops}, 2018.

\bibitem{karat2000}
J.~Karat, C.-m. Karat, and J.~Ukelson, ``Affordances, motivation, and the design of user interfaces,'' \emph{Commun. ACM}, vol.~43, pp. 49--51, 08 2000.

\bibitem{Masoudi_Fadel_Pagano_Elena_2019}
N.~Masoudi, G.~M. Fadel, C.~C. Pagano, and M.~V. Elena, ``A review of affordances and affordance-based design to address usability,'' \emph{Proceedings of the Design Society: International Conference on Engineering Design}, vol.~1, no.~1, p. 1353–1362, 2019.

\bibitem{affpersurvey}
\BIBentryALTinterwordspacing
M.~Hassanin, S.~Khan, and M.~Tahtali, ``Visual affordance and function understanding: A survey,'' \emph{ACM Comput. Surv.}, vol.~54, no.~3, Apr. 2021. [Online]. Available: \url{https://doi.org/10.1145/3446370}
\BIBentrySTDinterwordspacing

\bibitem{AfNet}
\BIBentryALTinterwordspacing
K.~M. Varadarajan and M.~Vincze, ``Afnet: The affordance network,'' in \emph{Asian Conference on Computer Vision}, 2012. [Online]. Available: \url{https://api.semanticscholar.org/CorpusID:39914795}
\BIBentrySTDinterwordspacing

\bibitem{AffordanceNet}
S.~Deng, X.~Xu, C.~Wu, K.~Chen, and K.~Jia, ``3d affordancenet: A benchmark for visual object affordance understanding,'' in \emph{Proceedings of the IEEE Conference on Computer Vision and Pattern Recognition}, 2021.

\bibitem{Nguyen2023open}
T.~Nguyen, M.~N. Vu, A.~Vuong, D.~Nguyen, T.~Vo, N.~Le, and A.~Nguyen, ``Open-vocabulary affordance detection in 3d point clouds,'' in \emph{Proceedings of the International Conference on Robotic Systems (IROS)}, 2023.

\bibitem{Luo2022LearningAG}
\BIBentryALTinterwordspacing
H.~Luo, W.~Zhai, J.~Zhang, Y.~Cao, and D.~Tao, ``Learning affordance grounding from exocentric images,'' \emph{2022 IEEE/CVF Conference on Computer Vision and Pattern Recognition (CVPR)}, pp. 2242--2251, 2022. [Online]. Available: \url{https://api.semanticscholar.org/CorpusID:247594536}
\BIBentrySTDinterwordspacing

\bibitem{Chen2021CerberusTJ}
\BIBentryALTinterwordspacing
X.~Chen, T.~Liu, H.~Zhao, G.~Zhou, and Y.-Q. Zhang, ``Cerberus transformer: Joint semantic, affordance and attribute parsing,'' \emph{2022 IEEE/CVF Conference on Computer Vision and Pattern Recognition (CVPR)}, pp. 19\,617--19\,626, 2021. [Online]. Available: \url{https://api.semanticscholar.org/CorpusID:244527378}
\BIBentrySTDinterwordspacing

\bibitem{Peng2022OpenScene3S}
\BIBentryALTinterwordspacing
S.~Peng, K.~Genova, ChiyuMaxJiang, A.~Tagliasacchi, M.~Pollefeys, and T.~A. Funkhouser, ``Openscene: 3d scene understanding with open vocabularies,'' \emph{2023 IEEE/CVF Conference on Computer Vision and Pattern Recognition (CVPR)}, pp. 815--824, 2022. [Online]. Available: \url{https://api.semanticscholar.org/CorpusID:254044069}
\BIBentrySTDinterwordspacing

\bibitem{mur2023multi}
L.~Mur-Labadia, J.~J. Guerrero, and R.~Martinez-Cantin, ``Multi-label affordance mapping from egocentric vision,'' in \emph{Proceedings of the IEEE/CVF International Conference on Computer Vision}, 2023, pp. 5238--5249.

\bibitem{luo2023learning}
H.~Luo, W.~Zhai, J.~Zhang, Y.~Cao, and D.~Tao, ``Learning visual affordance grounding from demonstration videos,'' \emph{IEEE Transactions on Neural Networks and Learning Systems}, 2023.

\bibitem{fang2018demo2vec}
K.~Fang, T.-L. Wu, D.~Yang, S.~Savarese, and J.~J. Lim, ``Demo2vec: Reasoning object affordances from online videos,'' in \emph{Proceedings of the IEEE Conference on Computer Vision and Pattern Recognition}, 2018, pp. 2139--2147.

\bibitem{damen2020epic}
D.~Damen, H.~Doughty, G.~M. Farinella, S.~Fidler, A.~Furnari, E.~Kazakos, D.~Moltisanti, J.~Munro, T.~Perrett, W.~Price \emph{et~al.}, ``The epic-kitchens dataset: Collection, challenges and baselines,'' \emph{IEEE Transactions on Pattern Analysis and Machine Intelligence}, vol.~43, no.~11, pp. 4125--4141, 2020.

\bibitem{grauman2022ego4d}
K.~Grauman, A.~Westbury, E.~Byrne, Z.~Chavis, A.~Furnari, R.~Girdhar, J.~Hamburger, H.~Jiang, M.~Liu, X.~Liu \emph{et~al.}, ``Ego4d: Around the world in 3,000 hours of egocentric video,'' in \emph{Proceedings of the IEEE/CVF Conference on Computer Vision and Pattern Recognition}, 2022, pp. 18\,995--19\,012.

\bibitem{bahl2023affordances}
S.~Bahl, R.~Mendonca, L.~Chen, U.~Jain, and D.~Pathak, ``Affordances from human videos as a versatile representation for robotics,'' in \emph{Proceedings of the IEEE/CVF Conference on Computer Vision and Pattern Recognition}, 2023, pp. 13\,778--13\,790.

\end{thebibliography}

\end{document}